\documentclass[10pt,twocolumn,letterpaper]{article}
\usepackage[pagenumbers]{wacv}

\usepackage{graphicx}
\usepackage{amsmath}
\usepackage{amssymb}
\usepackage{booktabs}

\usepackage{multirow}
\usepackage{pifont}
\newcommand{\xmark}{\ding{55}}

\usepackage{array}
\newcolumntype{P}[1]{>{\centering\arraybackslash}p{#1}}

\usepackage{xcolor}
\usepackage[pagebackref,breaklinks,colorlinks]{hyperref}

\newcommand{\name}{\texttt{AnyStar}}

\newcommand{\subpara}[1]{\vspace{0.4em} \noindent \textbf{#1}}

\usepackage[capitalize]{cleveref}
\crefname{section}{Sec.}{Secs.}
\Crefname{section}{Section}{Sections}
\Crefname{table}{Table}{Tables}
\crefname{table}{Tab.}{Tabs.}

\begin{document}

\title{\name: Domain randomized universal star-convex 3D instance segmentation}

\author{Neel Dey$^{1}$ { } { }
S.Mazdak Abulnaga$^{1}$ { } { }
Benjamin Billot$^{1}$ { } { }
Esra Abaci Turk$^{2}$ \\
P. Ellen Grant$^{2}$ { } { }
Adrian V. Dalca$^{1, 3}$ { } { }
Polina Golland$^{1}$\\
$^1$\small{MIT CSAIL} \hspace{0.75cm}
$^2$Boston Children's Hospital, Harvard Medical School  \hspace{0.75cm}
$^3$Martinos Center, Massachusetts General Hospital  \hspace{1cm}\\
}
\maketitle

\begin{abstract}
Star-convex shapes arise across bio-microscopy and radiology in the form of nuclei, nodules, metastases, and other units. Existing instance segmentation networks for such structures train on densely labeled instances for each dataset, which requires substantial and often impractical manual annotation effort. Further, significant reengineering or finetuning is needed when presented with new datasets and imaging modalities due to changes in contrast, shape, orientation, resolution, and density. We present \name, a domain-randomized generative model that simulates synthetic training data of blob-like objects with randomized appearance, environments, and imaging physics to train general-purpose star-convex instance segmentation networks. As a result, networks trained using our generative model do not require annotated images from unseen datasets. A single network trained on our synthesized data accurately 3D segments \textit{C. elegans} and \textit{P. dumerilii} nuclei in fluorescence microscopy, mouse cortical nuclei in $\mu$CT, zebrafish brain nuclei in EM, and placental cotyledons in human fetal MRI, all without any retraining, finetuning, transfer learning, or domain adaptation. Code is available at \url{https://github.com/neel-dey/AnyStar}.
\end{abstract}

\section{Introduction}

\paragraph{Motivation.} Assigning dense semantic labels for biomedical segmentation is difficult, individual instance annotations are even more expensive, and doing so in 3D is often infeasible. Even if a given dataset is painstakingly annotated, any subsequently trained segmentation network is unlikely to generalize to new datasets, scanners, and imaging configurations. For example, an instance segmentation network trained to segment \textit{C. elegans} nuclei in fluorescence microscopy is unlikely to also segment similarly-shaped nuclei in the mouse brain in micro-CT images. Consequently, high-throughput morphometric workflows across biosciences and radiology are bottlenecked by the need for reengineering and/or retraining networks for new datasets. Moreover, to retrain or adapt networks, biologists and clinicians need specialized hardware, data annotation pipelines, and machine learning expertise and infrastructure, which discourages rapid data analysis and adoption. We present a \textit{zero-shot} segmentation approach with appropriate appearance \& shape priors that addresses practitioner needs.

\begin{table}[!t]
\scriptsize
\caption{
Biomedical instance segmentation methods first train on real image-label pairs and then transfer to each new dataset with or without labels, corresponding to supervised/few-shot learning or domain adaptation, respectively. We instead synthesize image-label training data and generalize without labels or retraining.
}
\centering
\addtolength{\tabcolsep}{-0.325em}
\begin{tabular}{P{3.1cm}P{1.1cm}P{1.15cm}P{1.2cm}P{0.5cm}}
\toprule
\multirow{2}{*}{\textbf{Requirements}} & \textbf{Supervised Learning} & \textbf{Few-shot Prompting} & \textbf{Domain Adaptation} & \multirow{2}{*}{\textbf{Ours}} \\ \midrule
Real images for training data & \checkmark & \checkmark & \checkmark & \xmark \\
Manual labels for training data & \checkmark & \checkmark & \checkmark & \xmark \\
Manual labels for new datasets & \checkmark & \checkmark & \xmark & \xmark \\
(Re-)training on new datasets & \checkmark & \xmark & \checkmark & \xmark \\
\bottomrule
\end{tabular}
\label{table:setting}
\end{table}

\subpara{Current approaches.} Major efforts have been made to address these challenges. Domain adaptation methods train on annotated source domain images and either adapt trained networks to unlabeled target domain images~\cite{ouyang2019data} or perform cross-domain image translation~\cite{chen2020unsupervised,ren2021harmonization}. Unfortunately, these approaches typically succeed only when source and target domains are closely related~\cite{van2023unpaired}. Importantly, they also require biomedical specialists to train unstable and artifact-prone generative models (e.g. CycleGAN~\cite{zhu2017unpaired}) for each new dataset and to produce or acquire structurally similar annotated datasets to use as source domain data.

\begin{figure*}[!ht]
    \centering
    \includegraphics[width=0.98\textwidth]{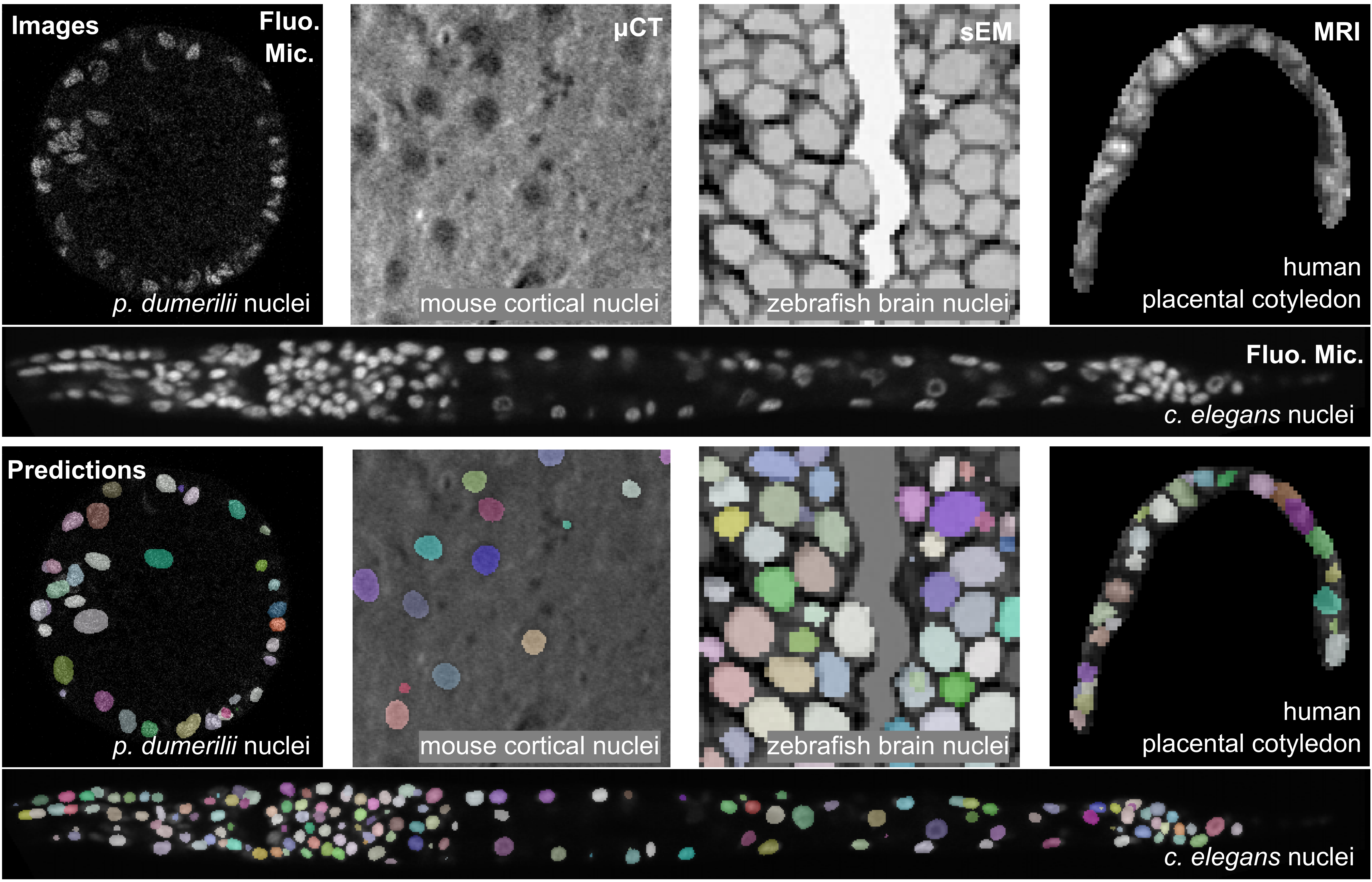}
    \caption{\textbf{Qualitative results.} After training an instance segmentation network only on 3D synthetic images generated by the \name~generative model, a \textit{single} trained network can segment foreground objects in real biomedical images across several imaging modalities, contrasts, and organisms without ever having seen any real images prior to testing and without any retraining or adaptation. 
    }
    \label{fig:highlight}
\end{figure*}

Another generalist approach collects large-scale inter-modality real-world annotated images for supervised training~\cite{pachitariu2022cellpose,stringer2021cellpose}. However, these methods require the new data to be well represented in the training corpus and are currently limited to 2D due to the difficulty of volumetric labeling. To our knowledge, there is no large-scale multi-organism and multi-modality corpus of 3D annotated instances to enable such generalist training with real data. 
Recent semantic segmentation methods in neuroimaging~\cite{billot2021synthseg,billot2022robust} employ domain randomization~\cite{tobin2017domain} and use training labels to simulate synthetic training images to generalize to new modalities and imaging configurations without finetuning in settings with relatively low variation such as anatomical neuroimage segmentation. In contrast, our goal of instance segmentation for several distinct biomedical foreground objects in arbitrary environments across organisms significantly expands image variability and necessitates novel methodology.

\subpara{Contributions.} We present \name, a generative model that synthesizes generalist training data to enable instance segmentation networks to segment any star-convex instance across bio-microscopy and radiology (Fig.~\ref{fig:highlight}). Biomedical targets such as nuclei and nodules can often be approximated by blob-like star-convex shapes~\cite{schmidt2018cell,weigert2020star}. Using this prior, we 
simulate image-label training data with domain-randomized object appearances, environments, densities, and imaging physics. 
An instance segmentation network trained using \name~gains empirical contrast-invariance in arbitrary environments. As a result, this network generalizes to five completely unseen datasets across biomedical microscopy and radiology without retraining. 
An \name-trained network approaches the segmentation accuracy of fully and weakly supervised domain-specific networks that require expensive annotations and outperforms supervised and/or pretrained networks when they are presented with out-of-domain data.
Finally, we investigate several generative modeling ablations with distinct intensity prior assumptions and find that a single generalist model that incorporates all assumptions is often sufficient.

\begin{figure*}[!ht]
    \centering
    \includegraphics[width=\textwidth]{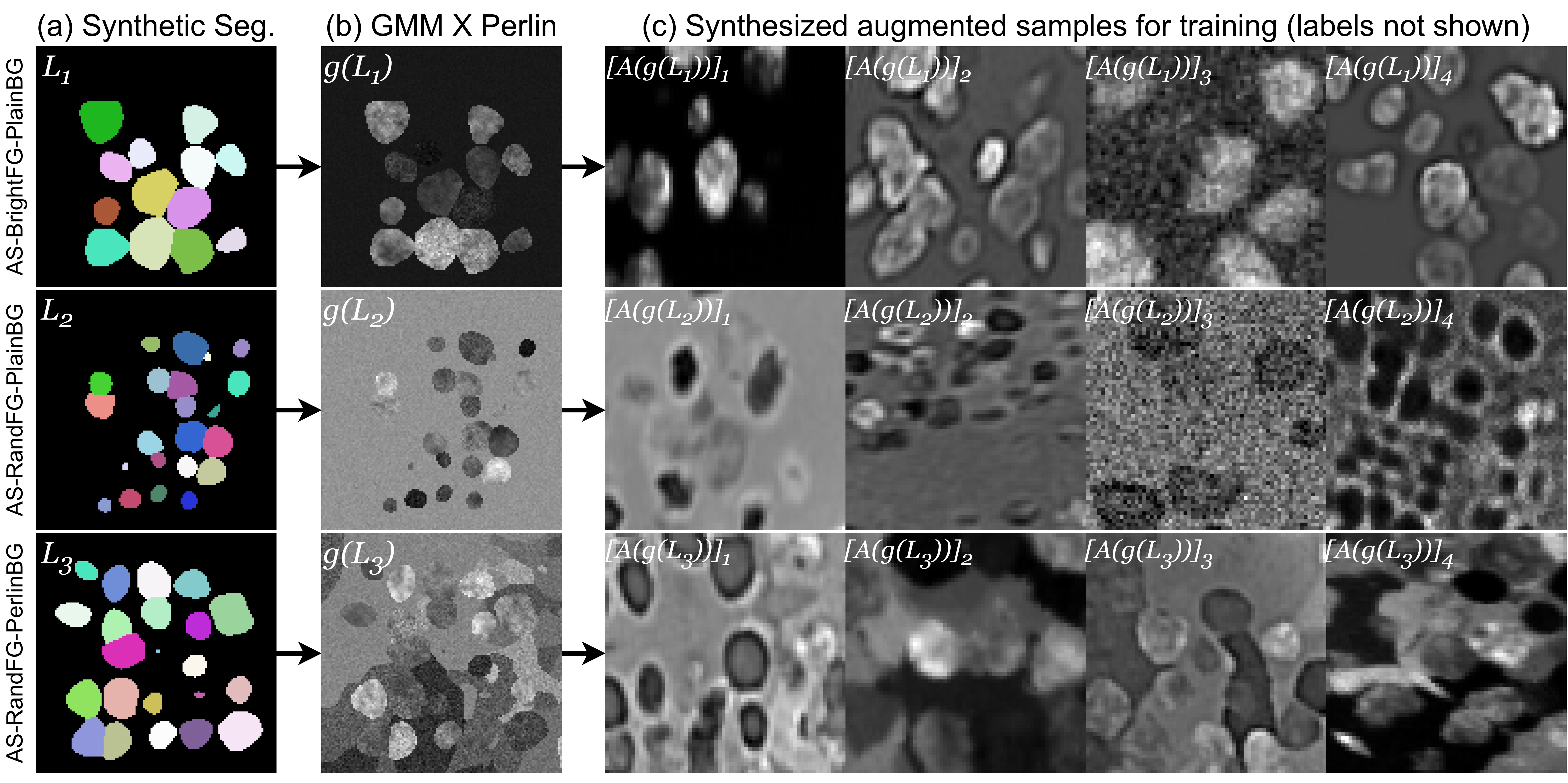}
    \caption{\textbf{Generative model.} Starting from $n$ labels in a synthetic segmentation $L$ (\textbf{col. 1}), we sample intensities in $g(L)$ from a $n$-component GMM which is pointwise modulated by Perlin noise (\textbf{col. 2}). A carefully designed augmentation sequence $A(\cdot)$ then simulates training data for instance segmentation (\textbf{cols. 3--6}). \textbf{Rows 1--3} showcase ablations with different specified priors over structure and contrast.}
    \label{fig:simulations}
\end{figure*}

\section{Related work}

\paragraph{Biomedical instance segmentation.} 
Established non-deep learning frameworks~\cite{carpenter2006cellprofiler,legland2016morpholibj,mcquin2018cellprofiler,sommer2011ilastik} typically use Otsu~\cite{otsu1979threshold}, feature engineering, and Watershed-based~\cite{beucher1982watersheds} pipelines to segment foreground objects. 
Using deep networks, early supervised semantic segmentation methods segmented cluttered objects by modeling a boundary class for improved separation~\cite{chen2016dcan,guerrero2018multiclass}. 
More recently, instance segmentation frameworks such as region-proposing~\cite{upschulte2022contour,zhao2018deep} and spatial embedding~\cite{lalit2022embedseg} networks typically achieve better instance-specific performance.
In particular,~\cite{schmidt2018cell,weigert2020star} found that fitting instances using star-convex shapes led to improved biomedical instance segmentation due to the wide applicability of the star-convex shape prior. 
Building on the star-convex shape prior, we develop a generative model that removes the need for retraining or adaptation and yields a universal star-convex instance segmentation network.

\subpara{Domain adaptation using generative models.} Given an annotated source dataset and a source-domain trained segmentor, cycle-consistent GAN losses are often used to close the domain gap to unlabeled target datasets. While successful in both microscopy~\cite{dunn2019deepsynth,lauenburg2022instance,wu2022nisnet3d,liu2020unsupervised} and radiology\cite{chen2020unsupervised,huo_synseg-net_2019,ren2021harmonization,chartsias2020disentangle}, these methods require domain pairs to contain similar structures for stable training and require retraining on each new dataset.
In contrast, our method does not require source data (whether annotated or not) for training as it does not need to adapt to new domains at test time.

\subpara{Generalist models and few-shot prompting.} 
Recently, large models pretrained on multiple real datasets have achieved strong segmentation performance on unseen natural and biomedical 2D datasets using test-time prompts like points/bounding boxes~\cite{kirillov2023segment} and context sets~\cite{butoi2023universeg}. 2D models such as Segment Anything~\cite{kirillov2023segment} can be applied zero-shot slice-wise as well to biomedical images but require finetuning and/or interactive filtering of extraneous predictions for usable performance\cite{ma2023segment,deng2023segment,huang2023segment}. Further, they are practically hard to finetune as 3D biomedical datasets with instance annotations are exceedingly rare and small in sample size, thus hindering their biomedical application. From another perspective, generalist bio-microscopy methods \cite{pachitariu2022cellpose,stringer2021cellpose} merge several 2D datasets to train models on diverse real images with the aim of generalizing to similar images. We instead perform 3D segmentation directly, focus only on star-convex objects, require no interaction or context sets, do not necessitate acquiring multiple real datasets, and generalize to structures not represented in multi-dataset corpora.

\begin{figure*}[!ht]
    \centering
    \includegraphics[width=0.98\textwidth]{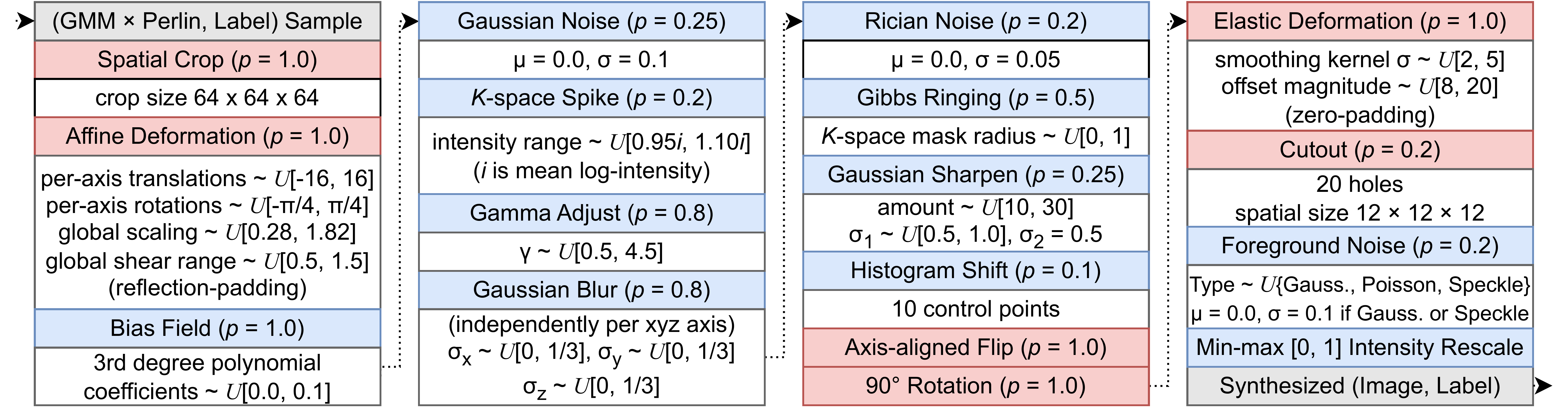}
    \caption{\textbf{Augmentation pipeline $A(\cdot)$.} \name~augmentations and their probabilities $p$ and hyperparameters (white boxes), read top-to-down left-to-right.
    Inputs and outputs are outlined in grey boxes and {\color[RGB]{214,132,154}joint image-label} and {\color[RGB]{108,142,191}image}-only augmentations are depicted by {\color[RGB]{214,132,154}red} and {\color[RGB]{108,142,191}blue} boxes, respectively. 
    $\mu$ and $\sigma$ denote means and standard deviations and other notational conventions follow~\cite{cardoso2022monai}.
    }
    \label{fig:augmentations}
\end{figure*}

\subpara{Realistic synthetic image generation.} Another class of unsupervised methods simulate label maps and train dataset-specific generative models to simulate training images~\cite{fu2018three,mahmood2019deep,dunn2019deepsynth,hou2019robust,eschweiler20213d,eschweiler2023denoising}.
These approaches typically train CycleGAN-like models to translate between the simulated organism-specific labels and real images from the target dataset. They then use the trained generative model to generate image-label pairs to train a second dataset-specific segmentation network. Our approach differs as it synthesizes both images and labels in a domain-agnostic manner and our generative model does not require any domain-specific training.

\subpara{Biomedical domain randomization.} Instead of synthesizing high-fidelity photorealistic training data, domain randomization methods~\cite{hoffmann2021synthmorph,tobin2017domain,tremblay2018training} anticipate domain shifts at deployment by synthesizing \textit{unrealistic} training examples with much higher variability using fully-controllable generative models. Consequently, domain-randomized semantic segmentation networks using a small number of training labels (without corresponding real images)~\cite{billot2022robust,billot2021synthseg} achieve strong generalization across modalities and resolutions in neuroimaging. In our work, alongside appearance randomization, \name~also performs label randomization to train a universal star-convex instance segmentor for disparate organisms in variable imaging environments.

\section{Methods}

\paragraph{Star-convexity.} A set $S \subset \mathbb{R}^{n}$ is star-convex if there exists an element $s_0 \in S$ that can be connected to each $s \in S$ by a line segment entirely within $S$. Following~\cite{weigert2020star}, we assume 3D biomedical foreground objects to be star-convex polyhedra parameterized by distances to the object boundary along pre-defined unit rays from each internal voxel.

\subpara{Label synthesis.} We first generate synthetic discrete foreground label maps (Fig.~\ref{fig:simulations}a) with oblong and irregularly shaped instances by adopting an existing synthetic nuclei generation approach~\cite{demodata_3D}. Specifically, we place $n$ spheres of radius $r$ and centers $c_i$, $i=\{1,\dots,n\}$ at the vertices of a regularly spaced 3D grid. The spheres are then independently randomly translated and scaled and up to a third of them are randomly removed. To simulate non-spherical shapes, the distances $d_{j}^i$ between each voxel $j$ (with coordinates $\mathbf x_j$) and object centers $c_1,\dots,c_n$ are corrupted by voxel-wise additive Perlin noise $p_j$~\cite{perlin1985image} as $d_{j}^i = \|\mathbf x_j - \mathbf c_i\|_{2} + 0.9rp_j$. Voxel $j$ is then assigned instance label $i$ if $\min_{i} d_j^i < r_i$ and is considered background otherwise. These initial label maps are zero- or reflection-padded independently along each axis to simulate varying instance densities and are scaled to a common image grid. 
We note that other label simulation approaches such as randomly distorted ellipsoids~\cite{svoboda2009generation} would yield visually similar images using our image synthesis pipeline described below.

\subpara{Intensity mixture modeling.} Given a label map $L$ with $n$ instances, we synthesize an initial image $g(L)$ (Fig.~\ref{fig:simulations}b). We sample foreground intensities of $g(L)$ from an $n$-component Gaussian mixture model (GMM) whose parameters $\{\mu_i, \sigma_i\}_{i=1}^n$ are drawn from a uniform distribution for each image. 
If a foreground voxel belongs to instance $i$, then its intensity is sampled from $\mathcal{N}(\mu_{i}, \sigma_{i}^{2})$.
We then apply multiplicative Perlin noise to emulate the spatial texture (e.g., staining differences) common in biomedical imaging.

\begin{table*}[!ht]
\footnotesize
\centering
\caption{\textbf{Experimental datasets.} 3D datasets used for evaluations. No real volumes are used for \name~training. Training data listed below are only used for training supervised baselines and validation data are used to tune probability and NMS thresholds only.
}
\centering
\begin{tabular}{ccccccccc}
\toprule
\textbf{Name} & \textbf{Organism} & \textbf{Modality} & \textbf{Image Resolution} & \textbf{Image Grid} & $\#_{\text{train}}$ & $\#_{\text{validation}}$ & $\#_{\text{test}}$ \\ \midrule
\texttt{CE}~\cite{hirsch2020auxiliary,long20093d,fuhui_long_2022_5942575} & \textit{C. elegans} & Fluo. Mic. & $0.116\times0.116\times0.122 \mu m^3$ & $1050\times140\times140$ & 18 & 3 & 7 \\
\texttt{NucMM-Z}~\cite{lin2021nucmm} & Zebrafish & sEM & $0.480\times0.510\times0.510 \mu m^3$ & $64 \times 64 \times 64$ & 25 & 2 & 27 \\
\texttt{NucMM-M}~\cite{lin2021nucmm} & Mouse & $\mu$CT & $0.480\times0.510\times0.510 \mu m^3$ & $192 \times 192 \times 192$ & 4 & - & 4 \\
\texttt{PlatyISH}~\cite{lalit2022embedseg} & \textit{P. dumerilii} & Fluo. Mic. & $0.450\times0.450\times0.450 \mu m^3$ & $300 \times 300 \times 300$ & 2 & - & 1 \\
\texttt{Placenta} & Human & BOLD & $3.000\times3.000\times3.000 mm^3$ & $80 \times 80 \times 64$ & \multicolumn{3}{c}{Qualitative evaluation} \\
\bottomrule
\end{tabular}
\label{table:datasets}
\end{table*}

\subpara{Background synthesis.} To model variable backgrounds and environments in $g(L)$, we investigate several choices corresponding to the rows of Fig.~\ref{fig:simulations}. To synthesize bright foreground instances, we model background intensities as an additional $(n+1)$th component in the GMM described above with $\mu_{n+1} < \min \{\mu_{1},\dots,\mu_{n}\}$. Alternatively, to simulate instances that may be brighter or darker than their surroundings, we simply sample $\mu_{1}, \dots, \mu_{n+1}$ uniformly at random as before. However, both assumptions do not yet account for instances that may be embedded in strongly textured environments with non-star-convex background structures as in radiology, for example. We therefore build on the \texttt{shapes} generative model of~\cite{hoffmann2021synthmorph} to simulate $b \sim \mathcal{U}\{1, B\}$ random geometric shapes in the background, again using a $b$-component GMM. To generate spatial background subcategories to later assign GMM components, we sample a $b$-channel Perlin noise volume and deform each channel independently with a smooth deformation. We next assign each background voxel a background sub-category corresponding to the channelwise argmax (not used during segmentation training). We then draw from a $b$-component GMM for each background voxel analogously to the foreground.

 \subpara{Ablations.} Our complete generative model, \textbf{\texttt{AS-Mix}}, uses all three of the above background models and randomly assigns one to each synthesized sample.
 Our generative model ablations include \textbf{\texttt{AS-BrightFG-PlainBG}} and \textbf{\texttt{AS-RandFG-PlainBG}} which simulate bright and randomized contrast foreground instances on untextured background, respectively.
 \textbf{\texttt{AS-RandFG-PerlinBG}} only uses textured backgrounds with randomized foreground contrast.

\subpara{Augmentation sequence.} $L$ and $g(L)$ are sampled to lie on a $128^3$ grid and are augmented by an extensive pipeline $A(\cdot)$ to generate the final training images as illustrated in Fig.~\ref{fig:simulations} cols. 3--6. We randomly crop $64^3$ subvolumes, followed by affine spatial deformations (translations, rotations, scales, and shears) with reflection padding used to simulate variable instance densities. We then employ several intensity augmentations including random bias fields, $k$-space spikes, Gibbs ringing, sharpening, gamma adjustments, and cutout\cite{devries2017improved} to simulate variable imaging artifacts.
Further, Gaussian blurring along each axis independently is used to simulate partial voluming common to anisotropic biomedical images.
This is followed by spatial deformations using random axis-aligned flips, 90$^{\circ}$ rotations, and elastic deformations with zero padding. Zero padding and cutout are used at the end of $A(\cdot)$ to simulate blank regions common in bioimaging. Finally, we add Gaussian, Poisson, or speckle noise to the non-zero regions. All augmentations and their stochastic probabilities are summarized in Fig.~\ref{fig:augmentations}.

\subpara{Segmentation network.} While the training data produced by our generative model can train any 2D or 3D instance segmentation network, we use a \texttt{StarDist} network~\cite{weigert2020star} as it matches our expected shape prior. \texttt{StarDist} regresses distance maps and ``centerness" probability maps whose dense label predictions are filtered by non-maximum suppression to obtain final segmentations. We use its default losses and hyperparameters with 96 rays.
As we focus on data generation, we train identical architectures and loss functions for all ablations and upper-bound supervised networks and change only their training and validation data.

\subpara{Implementation details.} The same 5-resolution U-Net architecture (with $2\times$ \texttt{Conv-BN-ReLU} blocks) is used for all methods, starting with 32 convolutional channels at the highest resolution and doubling thereafter after each max-pooling for all training runs. All networks are trained for 180,000 iterations using the Adam optimizer~\cite{kingma2014adam} with an initial learning rate of $2\times10^{-4}$ which is linearly decayed to 0. 
Fully and weakly-supervised baseline \texttt{StarDist} networks use the same architecture and optimization as \name, with their augmentations and training durations adjusted for their optimal performance. 
For fair comparison, we tune the object detection probability and non-maximum suppression thresholds of \texttt{StarDist} for all applicable methods on validation data as in~\cite{schmidt2018cell,weigert2020star}.
All \texttt{StarDist} networks are implemented using the open-source library: \url{https://github.com/stardist} and all augmentation implementations are taken from \texttt{MONAI}~\cite{cardoso2022monai}. 
Due to \name's extensive augmentation pipeline which can CPU-bottleneck training, we sample hundreds of thousands of augmented synthetic training volumes offline alongside  using inexpensive on-the-fly augmentations during training. Further implementation details are provided in the supplementary material.

\begin{figure*}[!ht]
    \centering
    \includegraphics[width=\textwidth]{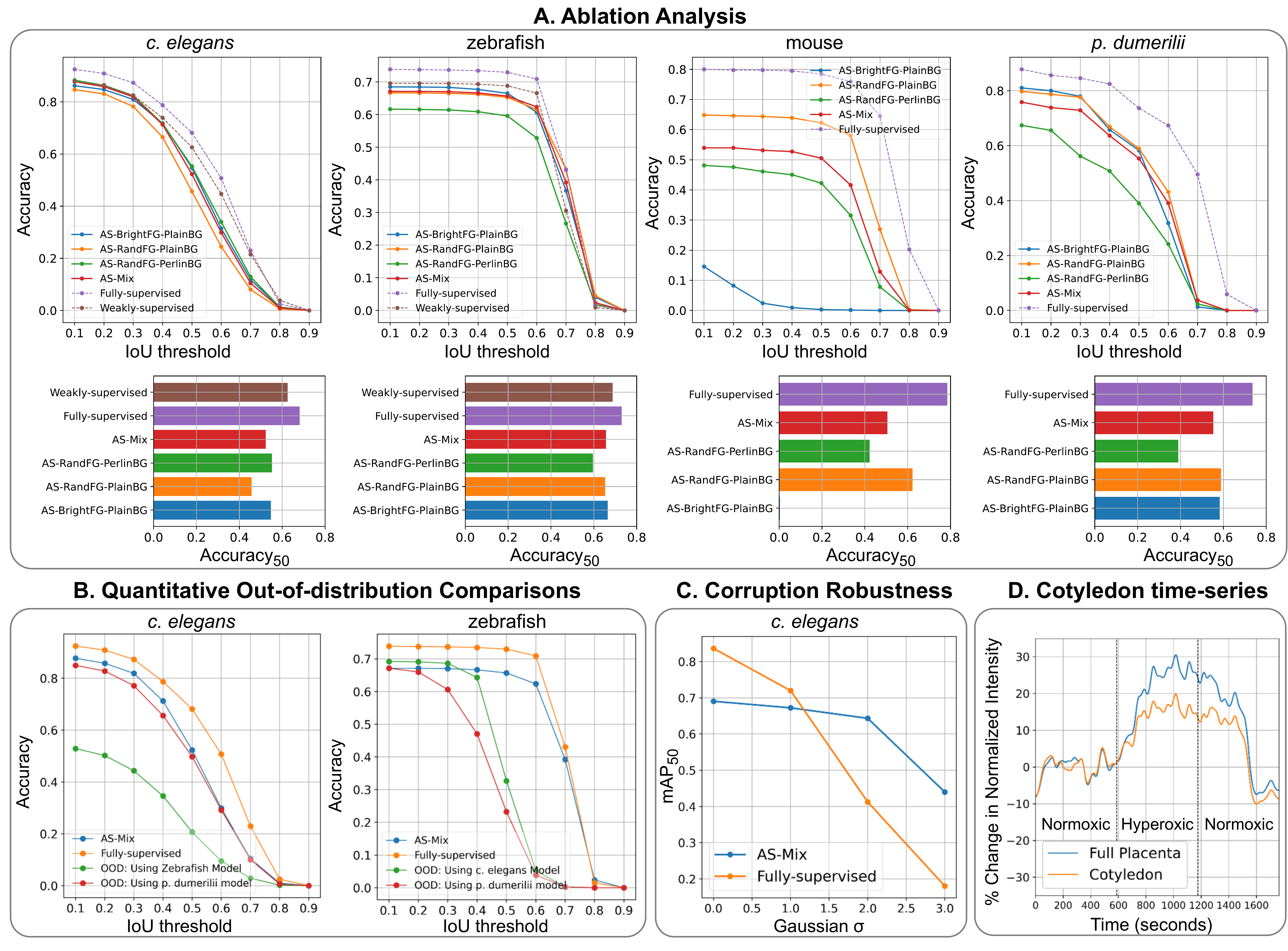}
    \caption{\textbf{Quantitative results.} \textbf{A.} \textit{Top:} Accuracy vs. IoU threshold analysis of the proposed ablations and target dataset-trained fully and weakly-supervised upper bounds across four 3D datasets. \textit{Bottom:} Accuracies at an IoU threshold of 0.5 for easier inspection.  \textbf{B.} Benchmarking zero-shot segmentation performance against fully-supervised networks trained on similar images from different but structurally-similar datasets. \textbf{C.} \name~networks are more stable against blur as opposed to domain-specific fully-supervised networks trained with blur augmentation. \textbf{D.} Zero-shot instance segmentation enables region-specific temporal analysis of placental MRI.}
    \label{fig:results_curves}
\end{figure*}

\section{Experiments}

\begin{figure*}[!ht]
    \centering
    \includegraphics[width=0.94\textwidth]{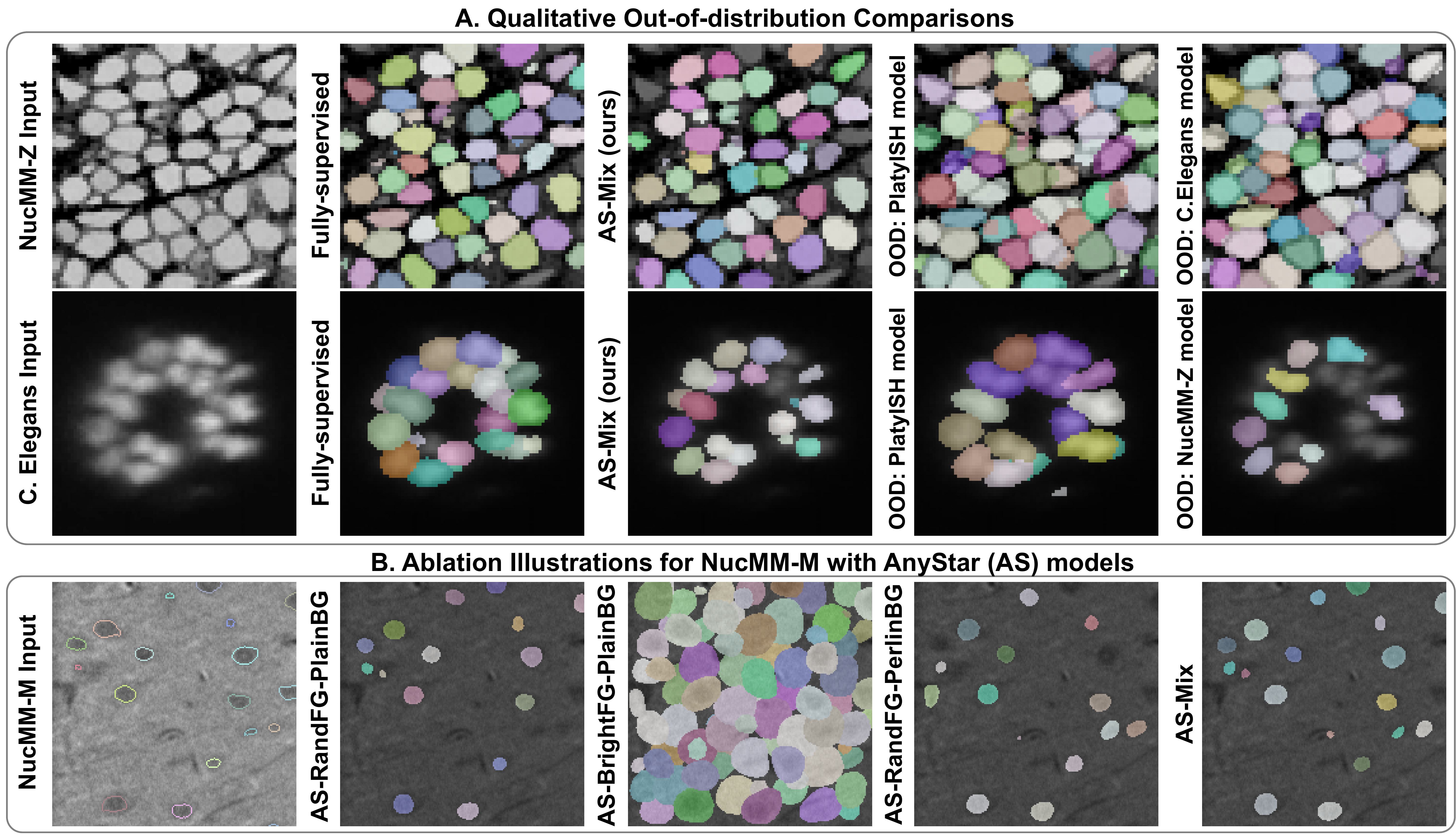}
    \caption{\textbf{Out-of-distribution and ablation result visualization}. \textbf{A.} Visualizations of zero-shot segmentation performance against fully-supervised networks trained on similar images from other domains. \textbf{B.} The outputs of ablation models on a mouse visual cortex sample (\texttt{NucMM-M}) with dark foreground (nuclei) contrast in $\mu$CT. Ablations using a narrow mis-specified intensity prior (\texttt{AS-BrightFG-PlainBG}) do not generalize across organisms and illustrate the importance of randomizing object contrast.
    }
    \label{fig:results_ood_ablation}
\end{figure*}

\paragraph{Data and preprocessing.} Table~\ref{table:datasets} summarizes the datasets and splits evaluated in this work. Our experimental data includes publicly available annotated datasets (\texttt{CE}, \texttt{NucMM}-\texttt{Z}~\&~\texttt{M}, \texttt{PlatyISH}) and a clinically-acquired fetal EPI MRI time-series dataset (\texttt{Placenta}) which is not annotated and thus qualitatively evaluated. These datasets are highly diverse and include a wide variety of object contrasts, densities, and background environments (Fig.~\ref{fig:highlight}, top). As \texttt{NucMM-M} and \texttt{PlatyISH} have limited samples, validation is performed on a held-out crop from the training set.

For the qualitative clinical application, we aim to segment individual placental \textit{cotyledons} in \texttt{Placenta} images as their MRI intensities are critical for characterizing fetal oxygenation~\cite{turk2019placental}. Therefore, non-placental tissue is removed from the images using a publicly-available network~\cite{abulnaga2022automatic} to focus on placental cotyledons. Further preprocessing details such as image registration, cropping, and resizing are provided in the supplementary material for all datasets.

\subpara{Baselines.} We establish \textit{upper-bounds} on performance using \texttt{StarDist} networks trained using all or one of the target dataset's training images, corresponding to fully- and weakly-supervised settings. 
Weakly-supervised networks are not trained for \texttt{NucMM-M} and \texttt{PlatyISH} datasets due to their limited sample sizes and supervised networks are not trained for \texttt{Placenta} as it does not have training labels.
We assess the \textit{out-of-domain} performance of these dataset-specific networks using the other datasets containing similar shapes and contrast.
As \name~does not use real images or annotations, its performance is expected to lie between out-of-domain and fully-supervised baselines. 

To test the performance of synthetic domain-randomized training data over collections of large-scale multi-dataset real images, we also compare against the generalist pretrained \texttt{cyto} and \texttt{nuclei} models available in CellPose~\cite{pachitariu2022cellpose}. These models are trained on large-scale publicly available real (non-synthetic) 2D biomedical microscopy images with manual annotations and are used as general-purpose microscopy image segmentors. As suggested, we use their slice-by-slice and along-all-planes 3D segmentation implementation in our experiments, tune the object diameter hyperparameter manually on validation data, and invert image contrast on the \texttt{NucMM-M} dataset where foreground units have dark contrast. We note that comparisons between our work and CellPose are confounded by changes in architecture and training loss.
Lastly, we exclude Segment Anything~\cite{kirillov2023segment} from comparisons as it is a 2D model, requires significant manual interaction in the form of points or bounding boxes for each 2D slice within large 3D biomedical volumes, and needs users to manually filter multiple extraneous region predictions in its zero-shot mode.

\subpara{Evaluation criteria.} All results are reported on held-out test data. To measure performance, we follow established instance segmentation evaluation strategies. A predicted instance is considered a true positive if it overlaps with a labeled instance by more than a specified IoU threshold. As in~\cite{lalit2022embedseg,weigert2020star,pachitariu2022cellpose,stringer2021cellpose}, to jointly measure detection and segmentation quality, we report mean accuracy
\footnote{mean average precision (mAP) is reported in the supplement.} 
computed against increasing the IoU threshold from 0.1 to 0.9 for in/out-of-domain supervised methods and our ablations. Further, as deployed networks may encounter unforeseen image corruptions, we measure the performance change of \texttt{AS-mix} and the fully-supervised network (trained with blur augmentation) when evaluated on images from the highest-resolution dataset (\texttt{CE}) with increasing strengths of Gaussian blurring. We report the mean average precision for these experiments at an IoU threshold of 0.5 with accuracy reported in the supplement. 
Lastly, previous work on placental oxygenation reports average BOLD intensity across time for the entire placenta which obfuscates differences between distinct subregions~\cite{sinding2018placental}. We speculate that cotyledon instance segmentation  may reveal biomedically-relevant cotyledon intensity time series within placentae.

\subsection{Results} 

\begin{figure*}[!ht]
    \centering
    \includegraphics[width=\textwidth]{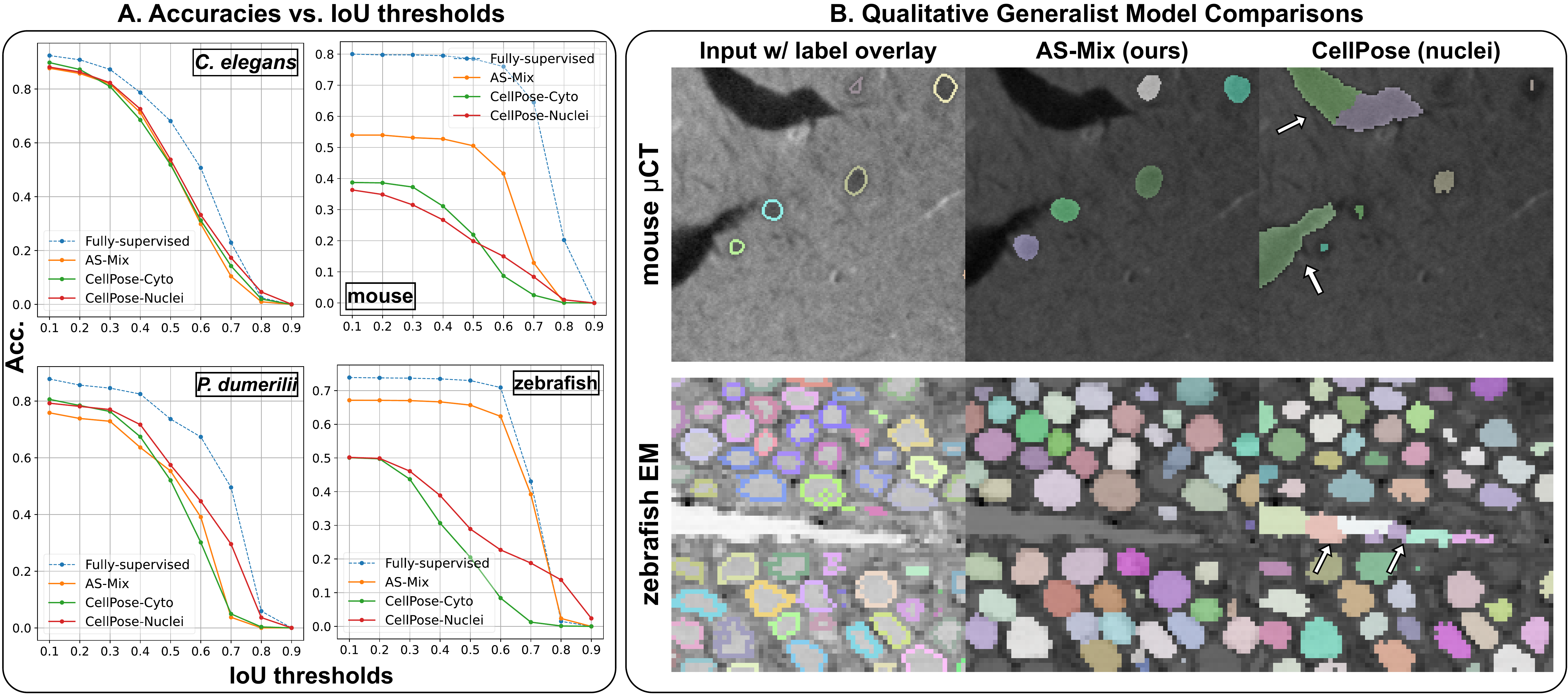}
    \caption{\textbf{Comparison against generalist models.} \textbf{A.} Accuracy vs. IoU threshold analysis of our model and two pretrained generalist CellPose models trained on a large multi-dataset corpus of real images~\cite{stringer2021cellpose,pachitariu2022cellpose}. \textbf{B.} Arbitrarily selected segmentation examples from the \texttt{NucMM-M} and \texttt{NucMM-Z} datasets produced by our model and CellPose. White arrows point to false positive predictions.}
    \label{fig:results_cellpose}
\end{figure*}

\noindent Fig.~\ref{fig:highlight} visualizes zero-shot segmentation predictions using \texttt{AS-mix} on all five datasets. Figs.~\ref{fig:results_curves} and~\ref{fig:results_cellpose}A report relevant quantitative statistics. We observe the following:

\subpara{\texttt{AS-mix} achieves strong inter-dataset generalization.} While a specific ablation may outperform others on a specific dataset, usually when its priors match the target dataset appearance, \texttt{AS-Mix} achieves consistently strong performance across all datasets (Fig. \ref{fig:results_curves}A). The \texttt{AS-BrightFG-PlainBG} ablation which trains only on images with bright foreground objects predictably fails when the target dataset (\texttt{NucMM-M}) has foreground instances with dark contrast as visualized in Fig.~\ref{fig:results_ood_ablation}B. Consequently, we recommend \texttt{AS-Mix} over other ablations for general use, unless the target dataset has a consistent intensity pattern matching one of our other ablations.

\subpara{\name~outperforms pretrained networks on unseen datasets.} Fully supervised methods provide an upper bound for in-domain performance when large sets of annotated images are available and task-specific models can be trained. Surprisingly, \name~approaches the performance of in-domain supervised methods without any in-task training data. Importantly, relative to supervised networks evaluated on unseen datasets containing instances of similar size, appearance, and contrast, \name~demonstrates consistently better performance (Figs.~\ref{fig:results_curves}B,~\ref{fig:results_ood_ablation}A). Representing a mild domain shift, a supervised network trained with extensive augmentation on \texttt{PlatyISH} nuclei in fluorescence microscopy does not generalize as well to \texttt{CE} nuclei as \name, which has never seen this modality. When evaluated on a larger domain gap via the \texttt{NucMM-Z} dataset, the fluorescence microscopy models trained on instances with similar contrasts underperform, whereas \name~produces better segmentations (Figs.~\ref{fig:results_curves}B, right;~\ref{fig:results_ood_ablation}A, top). 

In Fig.~\ref{fig:results_cellpose}, in comparison to CellPose's pretrained \texttt{cyto} and \texttt{nuclei} models, we find two distinct outcomes. On \texttt{CE} and \texttt{PlatyISH} whose imaging modality (fluorescence microscopy) and shapes (nuclei) are well represented in the CellPose training data, \texttt{AS-Mix} performs similarly to these pretrained models (Fig.~\ref{fig:results_cellpose}A, left col.). However, on datasets containing imaging modalities unseen by CellPose (\texttt{NucMM-Z}~\&~\texttt{M}), \texttt{AS-Mix} generalizes better, highlighting the benefit of training on randomized synthetic appearances.

\subpara{\name~gains robustness to blur degradation (Fig.~\ref{fig:results_curves}C).} Compared to a network trained with all available \texttt{CE} training data and augmentation (including blur), \name-\texttt{mix} better maintains segmentation performance as test images are progressively corrupted by Gaussian filtering, indicating improved robustness to texture changes.

\subpara{\name~enables exploratory data analysis on unlabeled data.} 
We select an arbitrary \texttt{Placenta} subject and compare the EPI time course over 321 motion-stabilized frames for the entire placenta vs. the average temporal intensities of the centroids of the segmented cotyledon regions. We smooth both their normalized intensities by a temporal Gaussian kernel with $\sigma=3.0$. In Fig.~\ref{fig:results_curves}D, using \texttt{AS-mix} we 
visualize the relative BOLD intensity changes in cotyledon subregions within the placenta with maternal oxygenation which showcases \name's practical utility on unlabeled datasets for exploratory data analysis. Further cotyledon segmentations are provided in the supplement.

\section{Discussion and conclusions}

\noindent\textbf{Limitations and future work.}
The presented method has limitations which will be addressed in future work.
For example, by definition, contrast invariant \name~networks will segment any star-convex object. This property may yield task-irrelevant predictions on datasets containing multi-contrast star-convex instances with only a few contrasts being of interest. However, task-specific prediction filtering methods based on intensity and shape priors would directly address these applications.
Also, as \name~does not currently represent non-star-convex objects such as neurons and vessels,  multiple other shape priors can be integrated into future pipelines.
Further, due to strong variability in biomedical image scales and the lack of a canonical resolution (as in neuroimaging~\cite{billot2021synthseg}), we use sliding window inference on large volumes that performs best when the patches contain objects that are sized similarly to the simulated objects. We expect that multi-scale training and inference methods would overcome this limitation. Importantly, dataset-specific (re-)training typically improves performance over zero-shot methods. If retraining expertise and infrastructure are available in biomedical centers, \name~can produce zero-shot segmentations that can be quickly refined to construct training sets, thereby reducing annotation effort and enabling rapid prototyping. 

\subpara{Conclusion.} 
In contrast to proposing a new architecture or loss, we focused on synthesizing data to train a generalist biomedical instance segmentation model.
To that end, we developed~\name, a domain-randomized generative model with a carefully designed stochastic appearance and shape model to simulate variable biomedical environments. A \textit{single} network trained on the synthesized data zero-shot segmented 3D biomedical objects across five unseen radiology and bioimaging datasets without any form of retraining and yielded both strong evaluation performance relative to current generalist approaches and enabled a novel and previously infeasible clinical application in fetal MRI.

\subpara{Acknowledgements.} We gratefully acknowledge funding from NIH NIBIB NAC P41EB015902, NIH NIBIB 5R01EB032708, NIH NICHD R01HD100009, NIH NIA  5R01AG064027, and NIH NIA 5R01AG070988. We thank Nalini M. Singh for helpful feedback and Martin Weigert for providing a reference OpenCL implementation of the \texttt{StarDist} demo nuclei simulator.

{\small
\bibliographystyle{ieee_fullname}
\bibliography{egbib}
}

\newpage
\clearpage
\appendix
\onecolumn

\section{Supplementary results}

\begin{figure*}[!h]
    \centering
    \includegraphics[width=\textwidth]{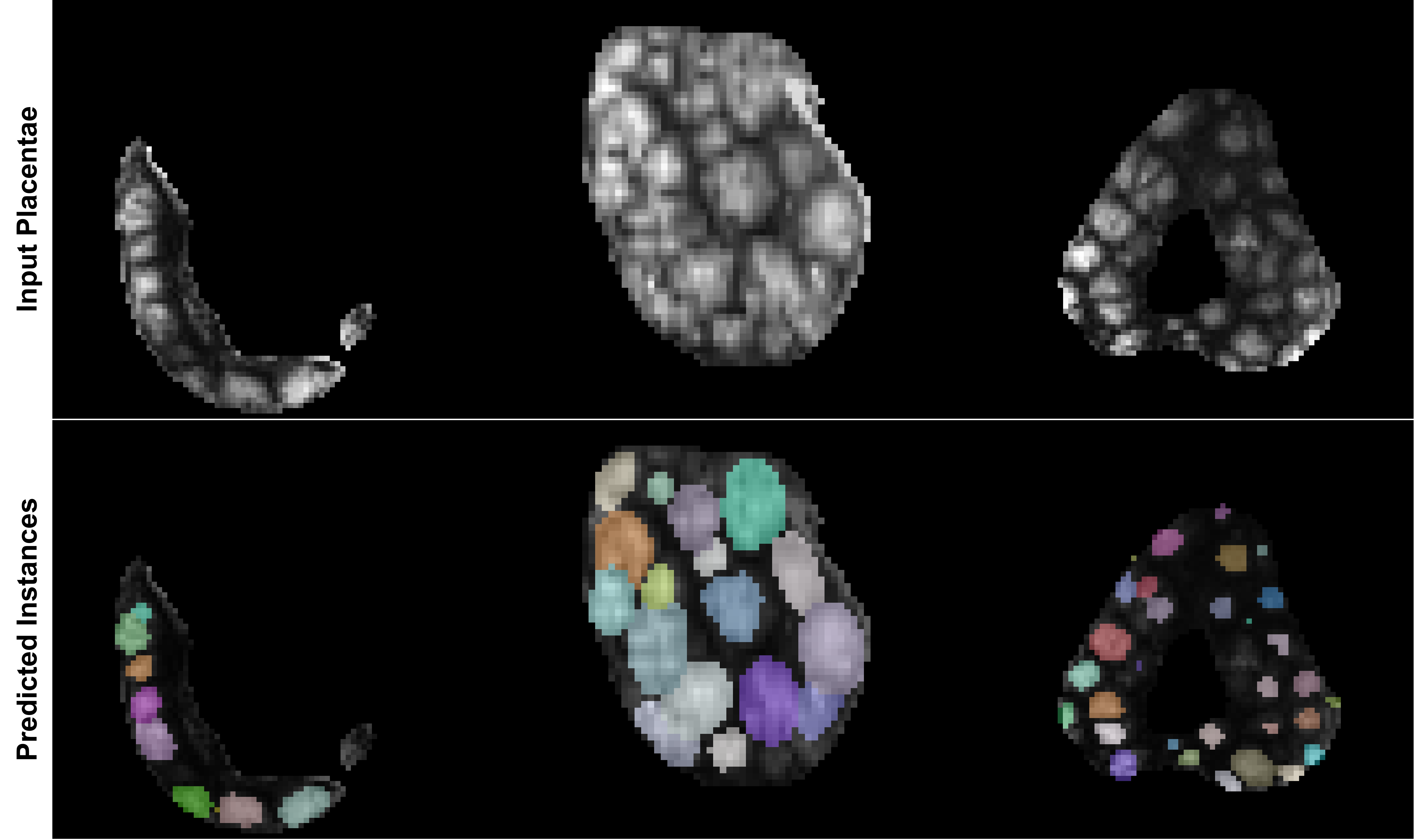}
    \caption{Qualitative 3D human placental cotyledon segmentations produced by \texttt{AnyStar-Mix}. Top: input image slices of 3D volumes, bottom: predicted objects. As ground-truth annotations are not available, we tune NMS and probability thresholds manually for qualitative visualization.}
    \label{fig:qualitative_coty}
\end{figure*}

\clearpage

\begin{figure*}[!ht]
    \centering
    \includegraphics[width=\textwidth]{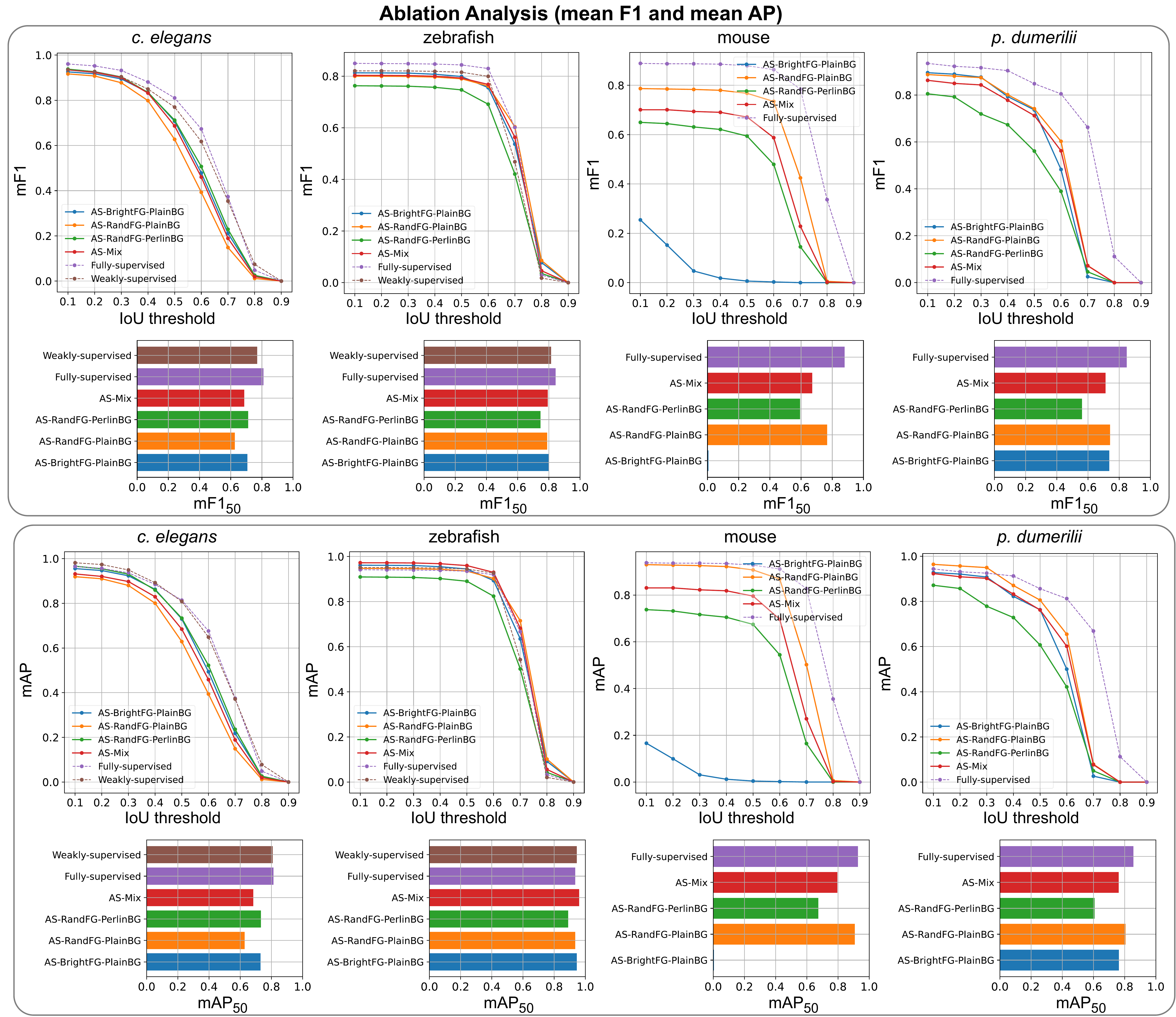}
    \caption{A companion figure to Figure 4 of the main text reporting both mean F1 (\textbf{top}) and mean average precision (\textbf{bottom})  vs. all IoU thresholds for our quantitative experiments, included for completeness.
    }
    \label{fig:suppl_ablation_scores}
\end{figure*}

\clearpage
\newpage 

\begin{figure*}[!ht]
    \centering
    \includegraphics[width=\textwidth]{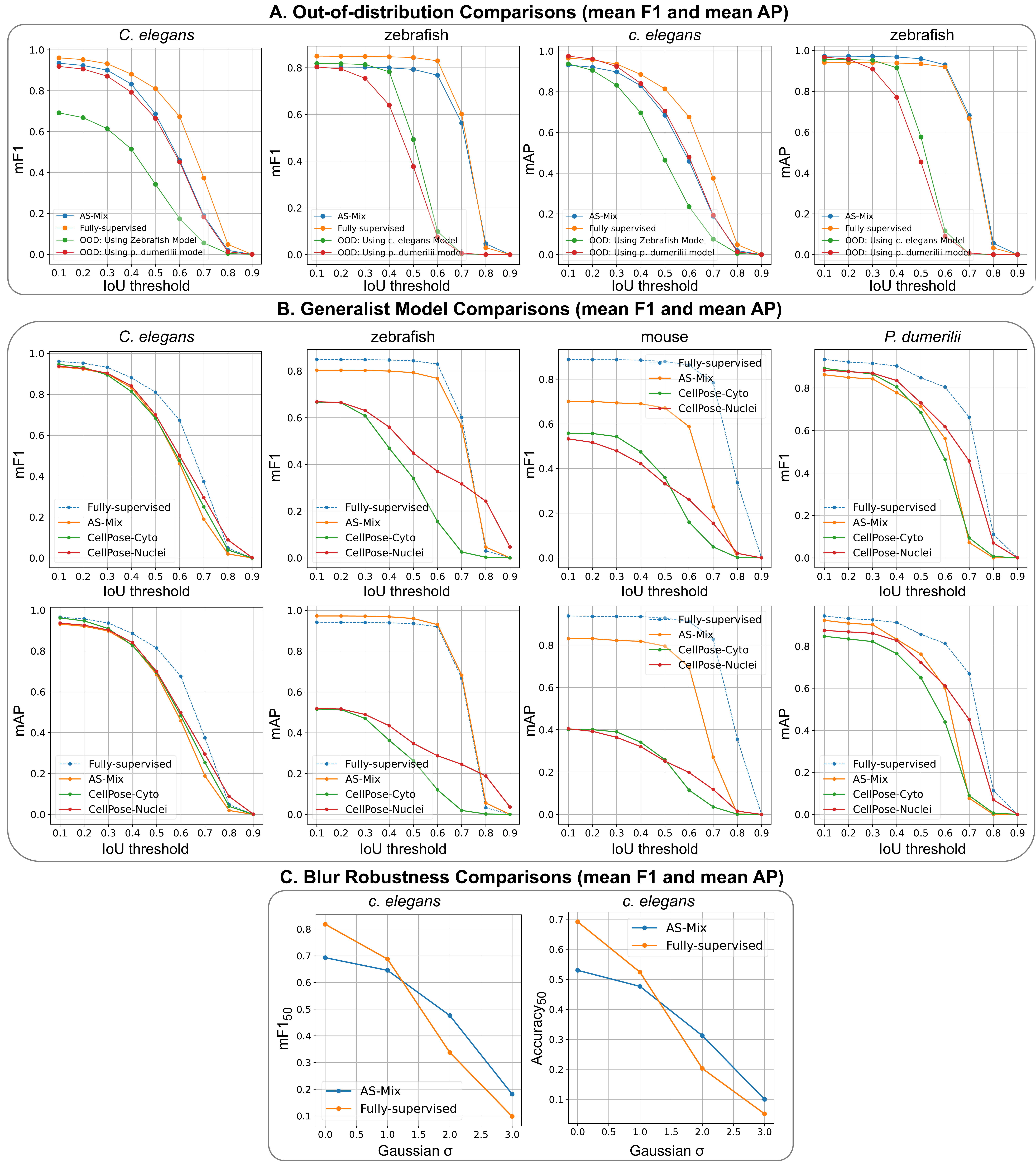}
    \caption{A companion figure to Figure 4 of the main text reporting both mean F1 and mean average precision vs. all IoU thresholds for our out-of-distribution (\textbf{A}), generalist model (\textbf{B}), and blur robustness (\textbf{bottom}) experiments, included for completeness.
    }
    \label{fig:suppl_ood_blur_scores}
\end{figure*}

\clearpage

\section{Additional experimental details}
\subsection{Data preparation}
\paragraph{Placenta.} Given a fetal BOLD MRI time-series from a subject, we first exclude non-placental tissue from analysis using a publicly-available segmentation network~\cite{abulnaga2022automatic}. We then motion-stabilize the temporal MRI sequence using the \texttt{ANTs} framework~\cite{tustison2020antsx} by jointly constructing an unbiased 3D template~\cite{joshi2004unbiased,avants2010optimal} and nonrigidly and diffeomorphically registering all temporal images to the template. Briefly, we use the local windowed NCC objective with a window size of 3 voxels, multi-scale registration, and B-Spline regularized SyN~\cite{tustison2013explicit} as a deformation model. Once stabilized, intensities can be sampled within placental subregions like cotyledons and are visualized in the main paper. As only qualitative segmentations are visualized and no supervised networks are trained, no data splitting is performed.

\paragraph{C. elegans.} The $yz$ plane images from~\cite{fuhui_long_2022_5942575} are cropped to a central 80 × 80 field-of-view and are resized to 64 × 64 along that plane for consistent processing across all methods. We use the dataset provided splits.

\paragraph{NucMM-Z \& M.} These datasets are available from~\cite{lin2021nucmm}. As \texttt{NucMM-Z} is already provided as annotated $64 \times 64 \times 64$ crops, we do not preprocess it in any way. \texttt{NucMM-M} is downsampled by a factor of 0.6 along all axes. As test sets for the two datasets are not publicly available, we use the provided original validation sets as test sets held-out for final evaluation. \texttt{NucMM-Z}'s original training set of 27 images is further split into a new 25/2 training/validation split used for early stopping.  As \texttt{NucMM-M} only has four training samples, we only use the training set for all model development and prototyping prior to final evaluation.

\paragraph{PlatyISH.} Lastly, due to the high (isotropic) resolution and low SNR of \texttt{PlatyISH}~\cite{lalit2022embedseg} samples, we foreground crop and downsample by a factor of $0.4$ along all axes as this was found to benefit methods using sliding window inference (\texttt{StarDist} networks and \texttt{CellPose}~\cite{stringer2021cellpose,pachitariu2022cellpose}) and to provide adequate denoising. We use the dataset provided splits. As \texttt{PlatyISH} only has two training samples, we only use the training set for all model development and prototyping prior to final evaluation.

\subsection{Other details}

\paragraph{Baseline augmentations.} Different augmentation pipelines are required to train on existing real data and obtain optimal performance as opposed to our approach of domain randomization which seeks to synthesize all forms of imaging artifacts and appearances from label maps. For example, real microscopy images do not typically have the MRI artifacts of bias fields, k-space spikes, Gibbs ringing, and cutout (MRI analysis often does organ-based masking). To that end, for fully and weakly supervised baselines trained on real microscopy images, we remove these MRI-specific transformations from their augmentation sequence. We retain randomized foreground cropping, gamma adjustments, blurring, histogram shifting, axis flips, 90-degree rotations, multi-distribution noise injection, and affine \& elastic deformations as real microscopy image augmentations.

\end{document}